\documentclass{ecai}
\usepackage{graphicx}
\usepackage{latexsym}

\usepackage{amsmath}
\usepackage{amsfonts}
\usepackage{amssymb}
\usepackage{algorithm}
\usepackage{algorithmic}

\usepackage{multirow}
\usepackage{hyperref}
\usepackage{booktabs}
\usepackage{colortbl}
\usepackage{svg}
\usepackage{subcaption}
\usepackage[normalem]{ulem}

\usepackage{CJKutf8}
\usepackage{color}
\usepackage[square,sort,comma,numbers]{natbib}

\ecaisubmission   

\begin{document}

\begin{frontmatter}

\title{Enhancing Text Generation with Cooperative Training}

\author[A]{Tong Wu\thanks{The work was done as an intern at IDEA.}$^{**;}$}\orcid{0009-0003-3154-1213}
\author[B]{Hao Wang\thanks{These authors contributed equally to this work.}}
\author[B]{Zhongshen Zeng}
\author[A]{Wei Wang}
\author[A,C]{Hai-Tao Zheng\thanks{Corresponding Authors. Email: zheng.haitao@sz.tsinghua.edu.cn, zhangjiaxing@idea.edu.cn}}
\author[B]{Jiaxing Zhang$^{***;}$}

\address[A]{Shezhen International Graduate School, Tsinghua Universiy}
\address[B]{International Digital Economy Academy}
\address[C]{Pengcheng Laboratory}



\begin{abstract}
Recently, there has been a surge in the use of generated data to enhance the performance of downstream models, largely due to the advancements in pre-trained language models. However, most prevailing methods trained generative and discriminative models in isolation, which left them unable to adapt to changes in each other. These approaches lead to generative models that are prone to deviating from the true data distribution and providing limited benefits to discriminative models. While some works have proposed jointly training generative and discriminative language models, their methods remain challenging due to the non-differentiable nature of discrete data. To overcome these issues, we introduce a \textit{self-consistent learning} framework in the text field that involves training a discriminator and  generator cooperatively in a closed-loop manner until a scoring consensus is reached. By learning directly from selected samples, our framework are able to mitigate training instabilities such as mode collapse and non-convergence. Extensive experiments on four downstream benchmarks, including AFQMC, CHIP-STS, QQP, and MRPC, demonstrate the efficacy of the proposed framework.
\end{abstract}

\end{frontmatter}

\begin{figure*}[ht]
\begin{center}
\includegraphics[width=0.8\textwidth]{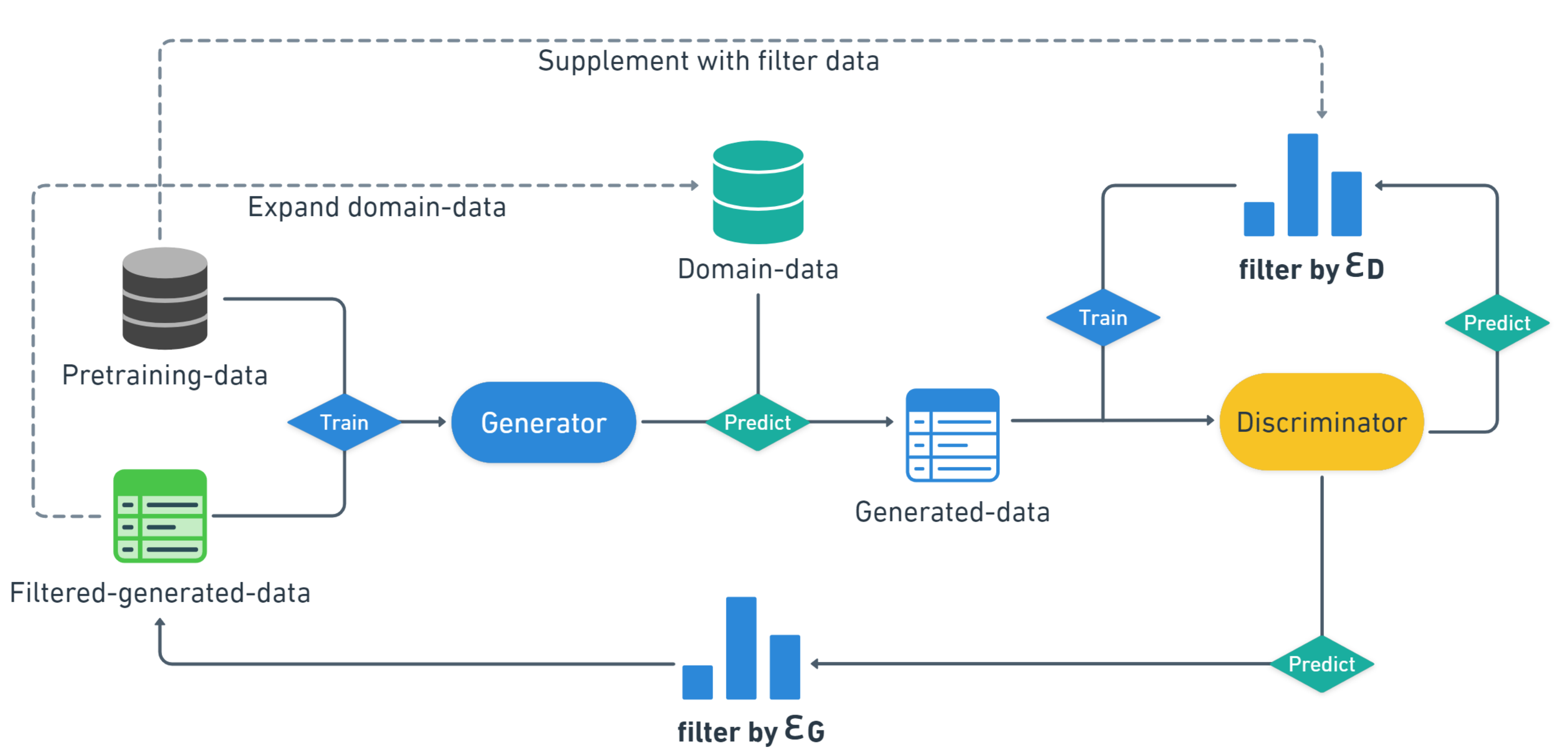}
\end{center}
\caption{Overview of the flow chart for the SCL framework.} 
\label{framework}
\end{figure*}

\section{Introduction}
The advance of Pre-trained Language Models (PLMs) like GPT-3 \cite{brown2020language} and LLaMA \cite{DBLP:journals/corr/abs-2302-13971} has substantially improved the performance of deep neural networks across a variety of Natural Language Processing (NLP) tasks. Various language models, based on the Transformer \cite{vaswani2017attention} architecture,  have been proposed, leading to state-of-the-art (SOTA) performance on the fundamental discrimination tasks. These models are first trained with self-supervised training objectives (e.g., predicting masked tokens according to surrounding tokens) on massive unlabeled text data, then fine-tuned on annotated data to adapt to downstream tasks of interest.  However, annotated data is usually limited to a wide range of downstream tasks, which results in overfitting and a lack of generalization to unseen data.

One straightforward way to deal with this data scarcity problem is data augmentation , and incorporating generative models to perform data augmentation has been widely adopted recently . Despite its popularity, the generated text can easily deviate from the real data distribution without exploiting any of the signals passed back from the discrimination task. In previous studies, generative data augmentation and discrimination have been well studied as separate problems, but it is less clear how these two can be leveraged in one framework and how their performances can be improved simultaneously. \looseness=-1

Generative Adversarial Networks (GANs) \cite{https://doi.org/10.48550/arxiv.1406.2661} are good attempts to couple generative and discriminative models in an adversarial manner, where a two-player minimax game between learners is carefully crafted. GANs have achieved tremendous success in domains such as image generation , and related studies have also shown their effectiveness in semi-supervised learning. However,  in the text field, GANs are difficult to train, most training objectives work well for only one model, either the discriminator or the generator, so rarely both learners can be optimal at the same time. This essentially arises from the adversarial nature of GANs, that during the process, optimizing one learner can easily destroy the learning ability of the other, making GANs fail to converge.

Another limitation of simultaneously optimizing the generator and the discriminator comes from the discrete nature of text in NLP, as no gradient propagation can be done from discriminators to generators. One theoretically sound attempt is to use reinforcement learning (RL), but the sparsity and the high variance of the rewards in NLP make the training particularly unstable \cite{caccia2019language}. 

To address these shortcomings, we novelly introduce a self-consistent learning framework based on one generator and one discriminator: the generator and the discriminator are alternately trained by way of cooperation instead of competition, and the selected samples are used as the medium to pass the feedback signal from the discriminator. Specifically, in each round of training, the samples generated by the generator are synthetically labeled by the discriminator, and then only part of them would be selected based on dynamic thresholds and used for the training of the discriminator and the generator in the next round. Several benefits can be discovered from this cooperative training process. First, a closed-loop form of cooperation can be established so that we can get the optimal generator and discriminator at the same time. Second, this framework helps improve the generation quality while ensuring the domain specificity of generator, which in turn contributes to training. Third, a steady stream of diverse synthetic samples can be added to the training in each round and lead to continuous improvement of the performance of all learners. Finally, we can start the training with only domain-related corpus and obtain strong results, while these data can be easily sampled with little cost or supervision. Also, the performance on labeled datasets can be further boosted based on the strong baselines. As an example to demonstrate the effectiveness of our framework in the text field, we examine it on four downstream text generation benchmarks, including AFQMC, CHIP-STS, QQP, and MRPC. The experiments show that our method significantly improves over standalone state-of-the-art discriminative models on zero-shot and full-data settings.

Our contributions are summarized as follows,

$\bullet$ We propose a self-consistent learning framework in the text field that incorporates the generator and the discriminator, in which both achieve remarkable performance gains simultaneously.

$\bullet$ We propose a dynamic selection mechanism such that cooperation between the generator and the discriminator drives the convergence to reach their scoring consensus.

$\bullet$ Experimental results show that the generator in our framework can continuously adjust its generation samples based on the performance of downstream tasks, while the discriminator can outperform the strong baselines.

\section{Related Works}
To alleviate the lack of annotated data in supervised learning in NLP,  semi-supervised learning (SSL) has been a popular line of research . The sources of the unlabeled data required by SSL are either collected from the domains or generated by generative language models. Then NLU models can learn from the unlabeled data by pseudo-labeling \cite{DBLP:conf/bmvc/Banitalebi-Dehkordi21} and consistent regularization \cite{sohn2020fixmatch}. However,  collecting unlabeled data comes at a cost(though smaller than labeling data), and the total amount is limited. Even with generative models, there is no guarantee of the quality of the generated samples, because the model cannot tune the generating results based on the performance of the downstream tasks. In contrast, our method usually includes a continuously updated generative model, which dynamically adjusts its generation according to the performance of downstream tasks.

GANs can be used as data enhancer to complement the lack of data for downstream tasks. Unlike conventional GANs in continuous domains, sequential GANs for discrete outputs are usually trained with reinforcement learning methods \cite{wu2021textgail}. But they usually suffer from high variance, partly due to the non-stationarity nature of their reward distribution. Whereas work based on cooperative training has opened the way for more efficient methods. CoT \cite{pmlr-v97-lu19d} explicitly estimates and optimizes JS divergence through a joint maximization framework, ConcreteGAN \cite{9296209} employs an autoencoder to learn implicit data manifold thus providing learning objective for adversarial training in a continuous space. However, their approach is still to back propagate through the gradient signal. More similar to our work is RML-GAN \cite{pmlr-v162-lamprier22a}, which uses a discriminator combined with a generative strategy to output real text samples for the task at hand. But they require complex and time-consuming Monte Carlo tree search, whereas we utilize a dynamic selection mechanism, and the training objective of the discriminator is exactly the same as that of the downstream task.

\section{Methodology}
\subsection{cooperative or adversarial}
Following the principle of self-consistency outlined in \cite{ma2022principles}, a closed-loop training needs to be built between the generator and the discriminator, either cooperatively or adversarially. GANs are typical examples of adversarial learning, but training  GANs remains quite unstable. Let us consider an extreme case to show the possible instability: the discriminator can perfectly distinguish real data and fake data generated by the generator, and the generator can fully reproduce the real data distribution. Then the discriminator has only a 50\% probability of selecting all samples that are generated by the generator. Therefore, any further updates to the generator parameters based on the feedback from the discriminator deviate the generator from the optimum. Neither the generator nor the discriminator can likely be optimal \cite{arjovsky2017towards}. In practice, a very delicate balance needs to be maintained between the discriminator and the generator to keep the training stable. In terms of cooperatively closed-loop learning, as discussed below, it does not suffer from instability: the generator and the discriminator usually enhance each other. 

\subsection{Self-consistent Learning Framework}
In this section, we introduce our self-consistent learning (\textbf{SCL}) framework. 

As shown in Figure~\ref{framework}, our framework, similar to the GANs, consists of a generator and a discriminator model. However, contrasting to the GANs, these two parts in our framework work cooperatively to enhance each other. Specifically, for any given class $k$, the generator $\mathcal{G}$ now become a conditional generator that takes in an input sentence $s^a_k$ and generate an output sentence $s^b_k$. The discriminator $\mathcal{D}$ is then responsible for discriminating the sentence using a dynamic threshold $\epsilon_\mathcal{D}$. The discriminated sentence is used as positive or negative data for that specific class to continue the training process. Once the new discriminator is trained, the sentence is discriminated again by the new discriminator with a different dynamic threshold $\epsilon_\mathcal{G}$. This time only the positive data is passed to the generator as the training data for the new round. In this way, a closed loop of cooperation is formed. \looseness=-1

In the above closed-loop training, we propose a \textbf{selection mechanism} that uses dynamic thresholds to filter samples. This mechanism is empirically shown to play a critical role in closing the gap between the generator and the discriminator, and thus makes this cooperation loop a virtuous circle. Specifically, as shown in Equation~\ref{mlp}, the output probability $p_{\mathcal{D}}(y=k|s^b_k)$ that the sentence $\{s^b_k\}$ belongs to class $k$ is calculated from the embedding representation $\mathbf{h}$\footnote{We follow \cite{reimers-gurevych-2019-sentence} and use the embedding representation of $CLS$-token as the sentence representation $\mathbf{h}$ .} of $\{s^b_k\}$,

\begin{equation}
\label{mlp} 
p_{\mathcal{D}}(y=k|s^b_k) = \text{softmax}( \text{MLP}(\mathbf{h}))
\end{equation}

where $y$ represents the class label.  Then, through the filtering function ${\texttt{filter}^{(t)}_k(\cdot)}$ in round $t$ for the $k$-th class 
 in Equation~\ref{thre}, we keep samples whose output probability is not less than threshold $\epsilon_{t,k}$, while other generated samples whose confidence is lower than threshold $\epsilon_{t,k}$ are discarded.

\begin{equation}
\label{thre}
    \texttt{filter}^{(t)}_k(s^b_k) \triangleq {p_{\mathcal{D}}(k|s^b_k) \geq \epsilon_k^t} 
\end{equation}

where $\epsilon_{t,k}$ represents the dynamic threshold for accepting $\{s^b_k\}$ as negative or positive samples in the $t$-th round. The generalized threshold function for $\epsilon_k^t$ is defined as,

\begin{equation}
\label{dyn_thre}
    \epsilon_k^t = f(t, \mathcal{L}_{t-1,k}, \epsilon_k^{t-1}) 
\end{equation}

where $\mathcal{L}_{t-1,k}$ and $\epsilon_k^{t-1}$ represent the discriminator loss and threshold for round $t-1$, respectively. $\mathcal{L}_{0,k}$ is set as 0 and $\epsilon_k^0 = \lambda$, where $\lambda$ represents a hyperparameter. \looseness=-1

\begin{theorem}
\label{theorem_kl}
At round $t$, given the previous round discriminator $\mathcal{D}^{t-1}_\phi$, the aim of the optimization of the generator $\mathcal{G}^{t}_\theta$,  boils down to, \looseness=-1
\begin{equation}
\nonumber
    \min_{\theta}  \mathbb{D}_{KL}(p_{\mathcal{D}^{t-1}_\phi}^k(\cdot), p_{\mathcal{G}^{t}_\theta}^k(\cdot)) 
\end{equation}

where $\mathbb{D}_{KL}$ is the standard KL divergence, $p_{\mathcal{G}^{t}_\theta}^k(\cdot)$ refers to the degree of confidence that the sentences generated by the generator belong to a given class $k$ (we can either train the generator to express its confidence in the generated sentences or use a fixed third-party model to score them), and $p_{\mathcal{D}^{t-1}_\phi}^k(\cdot)$ the probability of being classified into class $k$ given by the discriminator.
\end{theorem}

Theorem \ref{theorem_kl} shows that the generator at round $t$ is encouraged to approximate the probability distribution given by the previous round discriminator. In particular, on the basis of a well-pretrained discriminator, the generated distribution of the generator can be guaranteed to be faithful to the real data distribution. 

\textbf{Proof.} We use the previous round  generator $\mathcal{G}^{t-1}_\theta$ to generate samples, and filter them using the previous round discriminator $\mathcal{D}^{t-1}_\phi$ with a threshold $\epsilon^{t-1}$, then these samples are used for the training of the current round generator $\mathcal{G}^{t}_\theta$. Therefore, the optimization of $\mathcal{G}^{t}_\theta$ will tend to  maximize the probability that the generated samples pass the discrimination for the fixed $\mathcal{D}^{t-1}_\phi$. For a given class $k$, we have \looseness=-1
\begin{equation}
\nonumber
\small
 \max_{\theta} \mathbb{E}_{x\sim  p_{\mathcal{G}^{t-1}_\theta}^k} p_{\mathcal{G}^{t}_\theta}^k(x) \quad \texttt{s.t.} \quad \texttt{filter}^{(t-1)}_k(x) = 1
\end{equation}
where the definition of function $\texttt{filter}^{(t-1)}_k(\cdot)$ has been given in Equation~\ref{thre}.

The above objective is equivalent to sampling from the generator being optimized in round $t$ and making these samples pass the discrimination in round $t-1$ as much as possible, which gives
\begin{equation}
\nonumber
 \max_{\theta} \mathbb{E}_{x\sim  p_{\mathcal{G}^{t}_\theta}^k}p_{\mathcal{D}^{t-1}_{\phi}}^k(x)
\end{equation}
where $p_{\mathcal{D}^{t-1}_{\phi}}^k(x)$ is fixed.

A further transformation of the formula shows that
\begin{align*}
& \max_{\theta} \mathbb{E}_{x\sim  p_{\mathcal{G}^{t}_\theta}^k}p_{\mathcal{D}^{t-1}_{\phi}}^k(x) \\
\stackrel{(\romannumeral 1)} \Rightarrow & \max_\theta \int \mathrm{d}\pmb{\theta} \nabla_{\pmb{\theta}} \mathbb{E}_{x\sim  p_{\mathcal{G}^{t}_\theta}^k}p_{\mathcal{D}^{t-1}_\phi}^k(x) \\
\stackrel{(\romannumeral 2)} \Rightarrow
& \max_{\theta} \int \mathrm{d}\pmb{\theta} \mathbb{E}_{x\sim  p_{\mathcal{G}^{t}_\theta}^k} \nabla_{\pmb{\theta}} \log p_{\mathcal{G}^{t}_\theta}^k(x)p_{\mathcal{D}^{t-1}_\phi}^k(x) \\
\stackrel{(\romannumeral 3)} \Rightarrow
& \max_{\theta} \int \mathrm{d}\pmb{\theta}  \nabla_{\pmb{\theta}} \frac{1}{N}\sum_{i=1}^{N} \{\log p_{\mathcal{G}^{t}_\theta}^k(x_i)p_{\mathcal{D}^{t-1}_\phi}^k(x_i) \\&- \log p_{\mathcal{D}^{t-1}_\phi}^k(x_i)p_{\mathcal{D}^{t-1}_\phi}^k(x_i)\} \\
\stackrel{(\romannumeral 4)} \Rightarrow
& \min_{\theta} \mathbb{D}_{KL}(p_{\mathcal{D}^{t-1}_\phi}^k(\cdot), p_{\mathcal{G}^{t}_\theta}^k(\cdot))
\end{align*}

where $(\romannumeral 1)$ uses the integral property that integrating the derivative of a function gives the original function along with a constant, $(\romannumeral 2)$ takes advantage of the derivative property of the logarithmic function, $(\romannumeral 3)$ approximates the expectation of the probability distribution $p_{\mathcal{G}^{t}_\theta}^k(\cdot)$  by using averaging on $N$ samples sampling from $p_{\mathcal{G}^{t}_\theta}^k(\cdot)$, and adding a constant term $-\log p_{\mathcal{D}^{t-1}_\phi}^k(\cdot)p_{\mathcal{D}^{t-1}_\phi}^k(\cdot)$ with respect to $\theta$ under the summation would not change its derivative, and $(\romannumeral 4)$ cancels out the integral and the derivative and uses the definition of KL divergence. 
The above concludes our proof.

\textbf{Why Cooperative, Not Adversarial?} (1) the generator is no longer a challenger to the discriminator that only provides negative data points to fool it,  but now serves as a data augmenter to provide both positive and negative data points to enhance the discriminator; (2) the generator no longer updates its parameters through the policy gradients guided by the signals from the discriminator, but rather by utilizing the filtered data points to further improve its conditional generation quality. Note that by deliberately choosing the conditional generation paradigm along with the selection mechanism, we not only make the training more stable due to the different training goals, but also mitigate the mode collapse problem of GANs. Besides, by iterating through the loops, our framework achieves self-consistency by honing the domain specificity of the generator and increasing the domain data exposure of the discriminator.

\subsection{Text Generation}
We leverage the four text generation tasks ($i.e.$ $k=2$) as an example to demonstrate the effectiveness of our method. At this time, corresponding to Equation~\ref{thre}, $k=1/0$ represents the positive/negative class, and $\texttt{filter}^{(t)}_{1/0}$ represents the filter function in round $t$ for the positive/negative class respectively. First, let us introduce the formal definition of this task. Given two sentences ${s}^a = \{{w}_1^a, {w}_2^a, ..., {w}_{\ell_a}^a\}$ and ${s}^b = \{{w}_1^b, {w}_2^b, ..., {w}_{\ell_b}^b\}$, where ${w}_i^a$ and ${w}_j^b$ represent the $i$-th and $j$-th tokens in the sentences, and $\ell_a$ and $\ell_b$ indicate the  length of ${s}^a$ and ${s}^b$. The goal of this task is to learn a discriminator $\mathcal{D}$ to precisely predict the label $y=\mathcal{D}({s}^a, {s}^b)$, where $y \in \mathcal{Y}=\{0, 1\}$ indicates whether the two sentences are similar.

In our task, $\mathcal{G}$ is trained to generate a similar sentence ${s}^b$ from any given sentence ${s}^a$ and $\mathcal{D}$ is trained to predict label $y$ from any given sentence pair $\{{s}^a, {s}^b\}$. As demonstrated in Figure~\ref{framework}, there are mainly two training processes in the entire framework: fix $\mathcal{G}$ to train $\mathcal{D}$ and fix $\mathcal{D}$ to train $\mathcal{G}$. We introduce the two training procedures in detail with the $t$-th round training.

\textbf{Training $\mathcal{D}$}: We first randomly sample $s^a_t$ from domain-related corpus $C$, and then input $s^a_t$ to $\mathcal{G}^t$ to generate $s^b_t$. Next, we feed sentence pair $\{{s}^a_t, {s}^b_t\}$ into $\mathcal{D}^{t-1}$ to predict the label $y_{t-1}$, and filter $\{{s}^a_t, {s}^b_t, y_{t-1}\}$ using threshold $\epsilon_{\mathcal{D}}^{t-1}$. Finally, we train $\mathcal{D}^{t-1}$ on the selected data and pre-training data $P$ to get an improved discriminator $\mathcal{D}^{t}$. Note that the filtered data have both positive and negative samples. The update process of $\mathcal{D}$ seeks to minimize the cross-entropy loss over all instances:

\begin{equation}
\label{train_dis}
\begin{split}
	\mathcal{L}_\mathcal{D}(\boldsymbol{s}, \boldsymbol{y}) = \frac{1}{|\boldsymbol{s}|} \sum_{i=1}^{|\boldsymbol{s}|}-[y_i\cdot \log p_{\mathcal{D}}(y_i=1|s_i^a, s_i^b)\\+(1-y_i)\cdot \log (1-p_{\mathcal{D}}(y_i=1|s_i^a, s_i^b))] 
\end{split}
\end{equation}

\textbf{Training $\mathcal{G}$}: We feed the generated sentence pairs $\{{s}^a_t, {s}^b_t\}$ into $\mathcal{D}^{t}$ to predict new labels $y_{t}$, and then filter $\{{s}^a_t, {s}^b_t, y_{t}\}$ using threshold $\epsilon_\mathcal{G}^t$ and additional rules \footnote{The additional rules are used to exclude sentences which are too long, too short, or too similar according to the longest common substring algorithm.}. Note that the filtered data has only positive samples. For the filtered data, we supplement it with the pre-training data $P$ to update $\mathcal{G}^{t}$ to $\mathcal{G}^{t+1}$ \footnote{Note that the pre-training data $P$ is used to warm up $\mathcal{G}$ and $\mathcal{D}$. Although pre-training data is not mandatory in subsequent training, we empirically found that including it when training $\mathcal{G}$ can prevent language degeneration and improve downstream performances.} We also take out ${s}^b_t$ from the filtered data and add them to the domain-related corpus. The expanded domain corpus are used to sample conditional sentences in the next round of generation. The update procedure of $\mathcal{G}$ employs the negative log-likelihood function over all instances:

\begin{equation}
\nonumber
	\mathcal{L}_\mathcal{G}(\boldsymbol{s^a}, \boldsymbol{s^b}) = -\frac{1}{|\boldsymbol{s^b}|} \sum_{t=1}^{|\boldsymbol{s^b}|}\log p_{\mathcal{G}}(s^b_t|s^b_{<t}, \boldsymbol{s^a})
\end{equation}

For the selection mechanism, we adopt the form $\epsilon^t=m*t+\lambda$ after comparing the effects of different threshold functions through experiments according to Equation~\ref{dyn_thre}, where $m$ is the increment of the threshold for each round, $\lambda$ is the initial threshold, and $\epsilon^t$ is the threshold for rounds $t$.

In the process of training $\mathcal{G}$, since the sentences generated in each round are added to the domain-related corpus, the source of domain-specific data is thus monotonically expanding by iterating the self-consistent learning loop. The formalized process is shown in Algorithm~\ref{scl}. 


\begin{algorithm}
\caption{Self-consistent Learning (\textbf{SCL})} 
\label{scl}
\begin{algorithmic}[1]
\REQUIRE Generator $\mathcal{G}$; Discriminator $\mathcal{D}$; Domain-Related Corpus $C$; Pre-training Data $P$.
\STATE Initialize $\mathcal{G}^0$ and $\mathcal{D}^0$ with pre-trained language models;
\STATE Warm-up $\mathcal{G}^0$ and $\mathcal{D}^0$ with pre-training data $P$ to get $\mathcal{G}^1$ and $\mathcal{D}^1$;
\FOR {each round $i \in [1, n]$}
\IF {Two consecutive rounds of discriminator still improve}
\STATE Generate similar sentences $s^b\sim p_{\mathcal{G}^i}(\cdot|s^a)$ from sampled sentences $s^a$ from $C$;
\STATE Predict pseudo-labels $y^i \sim p_{\mathcal{D}^i}(\cdot|s^a, s^b)$;
\STATE Use threshold $\epsilon_\mathcal{D}^i$ to select data on $\{s^a, s^b, y^i\}$ to train $\mathcal{D}^{i+1}$;
\STATE Predict pseudo-labels $y^{i+1} \sim p_{\mathcal{D}^{i+1}}(\cdot|s^a, s^b)$;
\STATE Use threshold $\epsilon_\mathcal{G}^i$ and additional rules to select data on $\{s^a, s^b, y^{i+1}\}$ to train $\mathcal{G}^{i+1}$;
\ENDIF
\ENDFOR
\end{algorithmic}
\end{algorithm}

\section{Experiments}
\subsection{Tasks Design}
\label{task design}
In our experiments, the pre-training datasets are used to warm up the discriminator and generator, and the domain-related corpus is a set of independent sentences. To avoid label leakage, none of the training datasets participate in the pre-training of the generator and discriminator. In other words, the datasets in pre-training and self-consistent training are two non-overlapped datasets.

\textbf{Zero-Shot Baseline}: In the zero-shot setting, we utilize the warm-up generator and employ the constructed prompts to directly generate samples without any specific learning towards the prediction targets. These samples are then filtered by the discriminator and used as training data for the next round of the generator.We utilize the best-performing Chinese model RoBERTa-wwm-ext-large \cite{cui-etal-2020-revisiting}  and English model ALBERT-xxlarge-v2 \cite{Lan2020ALBERT:} as the base discriminators in our self-consistent learning framework. 

\textbf{Fine-Tune Baseline}: In the fine-tuning setting, similar sentence pairs like $<s_a, s_b>$ are used as training data for the generator in the form of "$s_a / s_b$ is similar to $s_b / s_a$". Here, "$s_a / s_b$ is similar to" serves as the prompt, and $s_b / s_a$ is the target that the generator will learn to predict. We compare our model with several strong baselines Chinese models MacBERT , StructBERT , RoFormer , XLNet, ELECTRA, ALBERT, RoBERTa and English models BERT, XLM-RoBERTa (XLM-R), XLNet, ELECTRA, ALBERT, RoBERTa.

\subsection{Experiments Setup}
\subsubsection{Datasets}
We conduct experiments on three Chinese datasets AFQMC (Financial) \cite{xu-etal-2020-clue}, CHIP-STS (Medical) \cite{zhang-etal-2022-cblue}, QQP-ZH (Common) \cite{wang-etal-2018-glue} and an English dataset MRPC (News) \cite{wang-etal-2018-glue}.  More details about the datasets are given in supplementary material.


\subsection{Zero-Shot Results}
\label{zeroshot res}
Table~\ref{dis_results}(a) shows how the F1 score of the discriminator varies with the number of self-consistent learning rounds on different datasets in the zero-shot task. According to Algorithm~\ref{scl}, the training is stopped when the discriminator no longer improves for two consecutive rounds. In addition, these four datasets are collected from different domains to further reflect the generality of our method in different domains.

\begin{table*}[t]
    \caption{Results of Cooperative Training through Self-Consistent Learning.} 
    \label{dis_results}
    \begin{subtable}{0.5\linewidth}
      \small
      \centering
        \caption{F1 Score of Discriminator in Zero-Shot Setting.}
        \begin{tabular}{c|cccc}
            \toprule
Round & AFQMC & CHIP-STS & QQP-ZH & MRPC \\ \midrule
0 & \uline{38.25} & \uline{58.82} & \uline{57.88} & \uline{68.54} \\ 
1 & 39.61 & 62.89 & 60.08 & 75.47 \\
2 & 44.98 & 67.24 & 58.57 & 76.63 \\
3 & 45.99 & 71.38 & 60.30 & 83.00 \\ 
4 & 45.71 & 71.45 & 61.31 & 83.90 \\ 
5 & 48.01 & 74.06 & 64.47 & 84.24 \\ 
6 & 50.41 & 74.08 & 66.44 & 84.50 \\ 
7 & 50.68 & 76.66 & 63.88 & 84.32 \\
8 & \textbf{51.36} & 76.30 & 65.46 & \textbf{84.61} \\ 
9 & - & 76.67 & 68.08 & - \\ 
10 & - & \textbf{77.42} & \textbf{70.51} & - \\ \midrule
 & \textbf{+13.11} & \textbf{+18.60} & \textbf{+12.63} & \textbf{+16.07}\\
            \bottomrule
        \end{tabular}
    \end{subtable}%
    \begin{subtable}{.5\linewidth}
      \centering
      \small
      \setlength\tabcolsep{3pt} 
      \sc
        \caption{F1 Score of Discriminator in Fine-Tune Setting.}
        \begin{tabular}{l|cccc}
            \toprule
Method & AFQMC & CHIP-STS & QQP-ZH & MRPC \\ \midrule
BERT\tiny{\textit{large}} & - & - & - & 82.51 \\ 
XLM-R\tiny{\textit{base}} & - & - & - & 84.27 \\  \midrule
MacBERT\tiny{\textit{large}} & 61.11 & 85.94 & 72.94 & - \\
StructBERT\tiny{\textit{large}} & 60.56 & 85.17 & 76.33 & - \\
RoFormer\tiny{\textit{large}} & \cellcolor[HTML]{C0C0C0}{64.19} & 84.16 & \cellcolor[HTML]{C0C0C0}76.56 & - \\  \midrule
XLNet\tiny{\textit{large}} & 50.31 & 82.97 & 64.96 & 79.51 \\
ELECTRA\tiny{\textit{large}} & 54.59 & 84.97 & 71.81 & 89.64 \\ 
ALBERT\tiny{\textit{large}} & 56.87 & 86.32 & 70.52 & \cellcolor[HTML]{C0C0C0}91.21  \\ 
RoBERTa\tiny{\textit{large}} & 57.29 & \cellcolor[HTML]{C0C0C0}86.93 & 74.58 & 90.24 \\ 
\midrule
Self-Consistent & \textbf{66.59} & \textbf{88.39} & \textbf{78.43} & \textbf{92.78} \\ 
            \bottomrule
        \end{tabular}
    \end{subtable} 
\end{table*}

The scores in the last line of Table~\ref{dis_results}(a) give the improvement of our discriminator in the last round relative to the first round. We can see that the F1 score gradually increases after each training round, eventually reaching a 10+ absolute percentage (AP) improvement. We believe what drives the improvement of the discriminator is the self-consistency, which it acquires with the generator step by step during the loop. 

To verify that the generator also improves after self-consistent training, we adopt Perplexity and Bertscore to measure the language fluency and the semantic similarity (i.e. domain specificity) respectively. For different generators in different rounds, we first select $s^a$ in similar sentence pairs from the same test set as the original sentences input, and generate similar sentences $s^b$ with greedy search. The reason for not using other sampling methods is to ensure reproducibility. Given the generated sentences, we introduce an additional GPT2 \footnote{Wenzhong-GPT2-110M for Chinese data, and GPT2-base for English data.} model to calculate the perplexity of generated similar sentences, and use a third-party library \footnote{\href{https://pypi.org/project/bert-score/}{https://pypi.org/project/bert-score/}} to calculate the bertscore between the original and generated similar sentences. The results are shown in Table~\ref{gen_res}. 

\begin{table}
\small
\centering
\setlength\tabcolsep{6pt} 
\caption{Zero-Shot Performance of Generator in Zero-Shot Setting.}
\label{gen_res}
\begin{tabular}{l|cccc}
\toprule
 &  AFQMC & CHIP-STS & \begin{tabular}[c]{@{}c@{}}QQP-ZH \end{tabular} & MRPC \\ \midrule
\begin{tabular}[c]{@{}r@{}}Perplexity ↓\\\textit{-first round}\end{tabular} & 10.13 & 6.86 & 12.94 & 28.71 \\
\begin{tabular}[c]{@{}r@{}}Perplexity ↓\\\textit{-last round}\end{tabular} & 8.43 & 5.97 & 12.27 & 17.56 \\ \midrule
\begin{tabular}[c]{@{}r@{}}Bertscore ↑\\\textit{-first round}\end{tabular} & 0.79 & 0.84 & 0.87 & 0.94 \\ 
\begin{tabular}[c]{@{}r@{}}Bertscore ↑\\\textit{-last round}\end{tabular} & 0.80 & 0.85 & 0.89 & 0.97 \\ \bottomrule
\end{tabular}
\end{table}

We can see that the perplexity / bertscore of the last round in Table~\ref{gen_res} has decreased / improved compared to the first round. Note that a lower perplexity indicates a more fluent sentence, while a higher bertscore indicates a more similar sentence. It suggests that after self-consistent training,  the generator is gradually improved in language fluency and semantic similarity (i.e. domain specificity). The reason why the improvement of the generator is not as obvious as that of the discriminator is that the size of the generator is several times that of the discriminator, and the total number of training samples is limited. In supplementary material, the generated samples of the generator in different rounds are given to show the changes in the generation.

\subsection{Fine-Tune Results}
Our method not only works well in the zero-shot case, but also achieves good results in the full-data case. For the sake of a fair comparison, we reproduce several strong baselines on the four training sets, and their performances on the test sets are shown in Table~\ref{dis_results}(b).

Our approach uses the best-performing model on a single test set as the base discriminator for self-consistent learning. The bold scores in the last line of Table~\ref{dis_results}(b) show that our method outperforms the strong baselines (shaded in gray) by 1 to 2 AP on all four test datasets, indicating the potential of self-consistent learning to further improve the model performance. 

\subsection{Evaluating Self-consistency}
In this section, we evaluate the consistency between the generator and the discriminator as the learning loop unfolds. We follow the same method used in Section~\ref{zeroshot res} and use greedy search to generate similar sentences on the same test set. Then we take the confidence of the discriminator $R_\mathcal{D}$ as the score of the discriminator, which is calculated for the original sentences $s^a$ and the generated similar sentences $s^b$ according to Equation~\ref{r_d}. 

\begin{equation}
	R_\mathcal{D} = p_\mathcal{D}(y^+|s^a, s^b) \label{r_d} 
\end{equation}

where $y^+$ represents a positive label. 

\begin{figure*}[ht] 
\begin{center}
\includegraphics[width=0.8\textwidth]{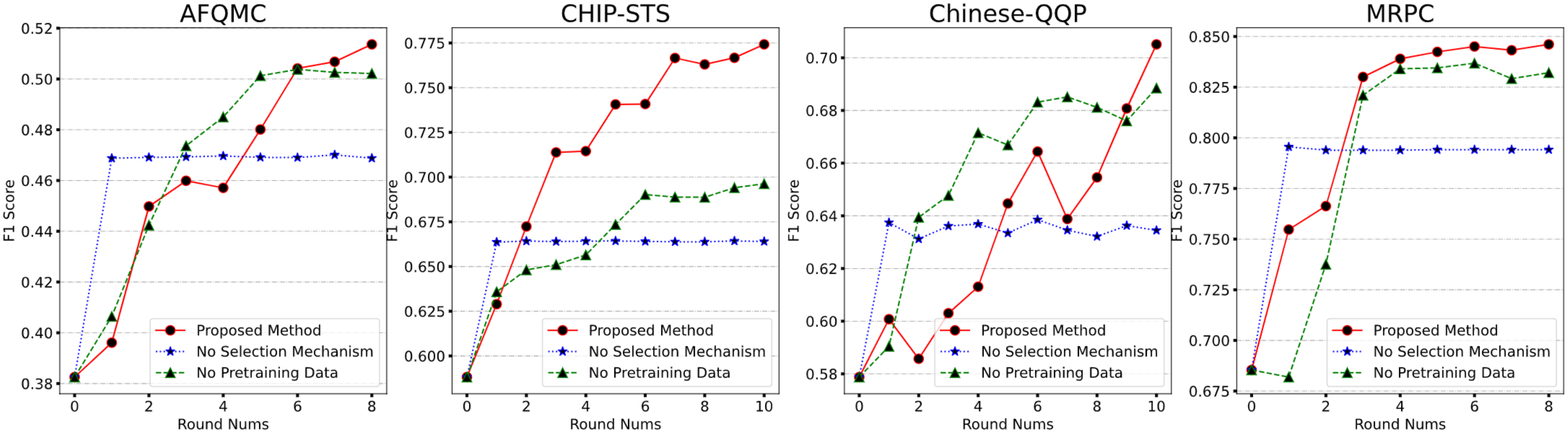}
\end{center}
\captionsetup{width=0.9\textwidth}
\caption{Results of ablation experiments on pre-training data and selection mechanism of Zero-Shot. Results of the proposed method, without pre-training data, and without the selection mechanism are given in red, green, and blue, respectively.} 
\label{ablation}
\end{figure*}

For the generator, using its own perplexity as a criterion for determining the similarity between sentences $s^a$ and $s^b$ is not always effective. Perplexity primarily reflects the generator's ability to fit similar data pairs, but it falls short in mitigating the impact of noise pairs. Therefore, to quantify this similarity, we introduce a third-party static model SimCSE \footnote{We use SimCSE-BERT-base to calculate scores on Chinese datasets and sup-SimCSE-BERT-base-uncased on English datasets. } to get the embedding representation $\mathbf{a},\mathbf{b}$ of sentences $s^a,s^b$. The cosine similarity $R_\mathcal{G}$ between $\mathbf{a}$ and $\mathbf{b}$ is then calculated according to Equation~\ref{r_g} to approximate the score of the generator.

\begin{gather}
    \mathbf{a}, \mathbf{b} = \text{Encoder}(s^a), \text{Encoder}(s^b) \nonumber \\ 
	R_\mathcal{G} = \frac{\mathbf{a} \cdot \mathbf{b}}{\left\|\mathbf{a}\right\|_2*\left\|\mathbf{b}\right\|_2} 
\label{r_g}
\end{gather}

where $\mathbf{a}$ and $\mathbf{b}$ both represent the embedding representation at the $[CLS]$ position. Note that the original sentence $s^a$ remains unchanged in each round, while the generated sentence $s^b$ changes.

\begin{figure}[ht] 
\begin{center}
\includegraphics[width=0.3\textwidth]{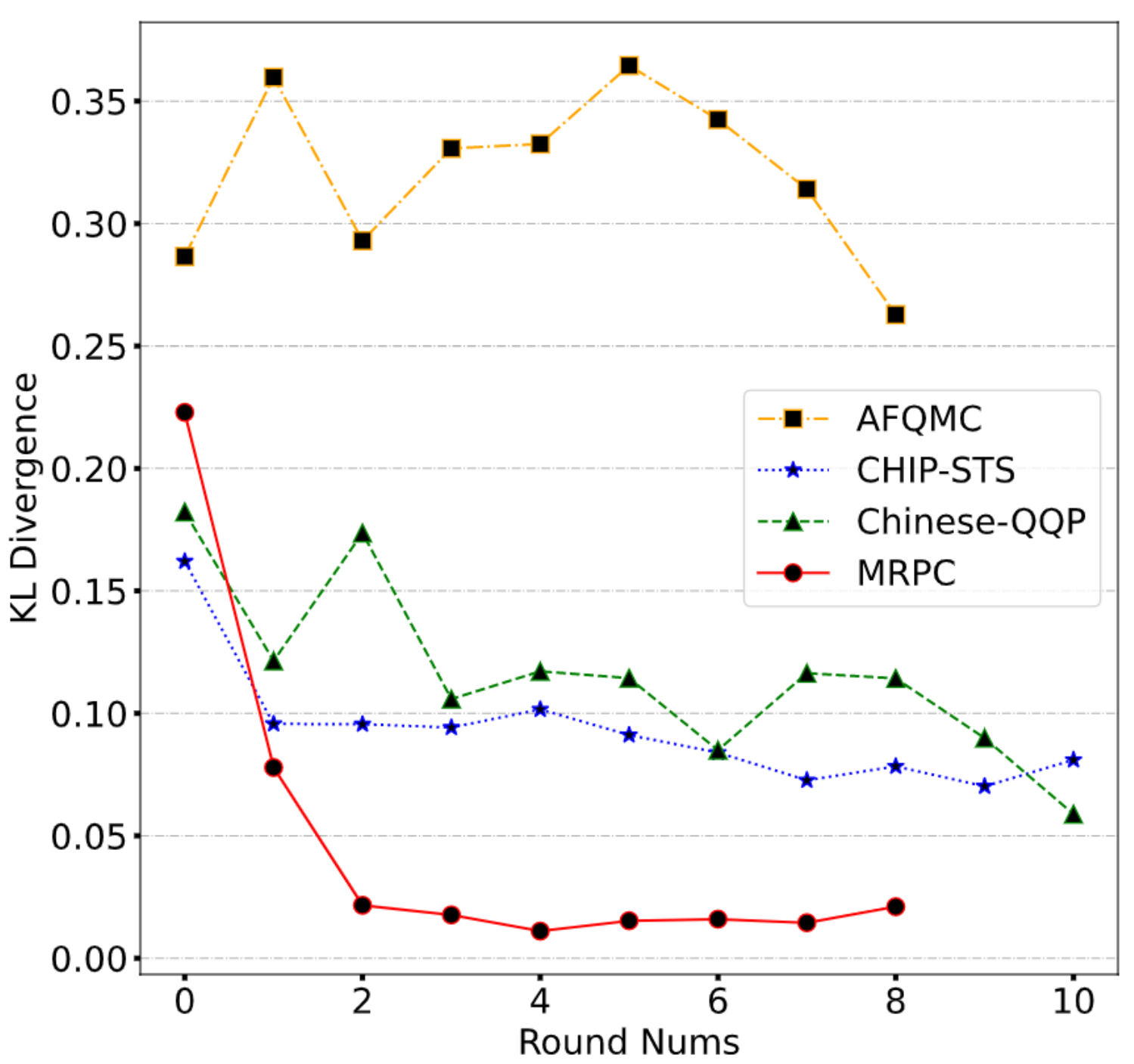}
\captionsetup{width=0.4\textwidth}
\caption{The KL Divergence between the score distributions of Discriminator and Generator in Zero-Shot.} 
\label{distance}
\end{center}
\end{figure}

Finally, for the trained discriminator and generator in each round $t$, we can obtain two score distributions $\mathbf{R_\mathcal{D}^t}$ and $\mathbf{R_\mathcal{G}^t}$ correspondingly. According to Theorem~\ref{theorem_kl}, we draw the curves of KL divergence between $\mathbf{R_\mathcal{D}^t}$ and $\mathbf{R_\mathcal{G}^t}$ in each round for the four datasets: AFQMC, CHIP-STS, QQP-ZH, and MRPC. As illustrated in Figure~\ref{distance}, all the curves show a clear downward trend, indicating that the distance between the two score distributions decreases with the increase in the number of training rounds until a score consensus is reached.

\subsection{Effect of Pre-training Data and Selection Mechanism}
\label{ablation exp}
We perform ablation experiments on the pre-training data and the selection mechanism in the zero-shot case. As described in Section~\ref{task design}, the pre-training data is used to pre-train the generator and discriminator, completely independent of the experimental datasets in self-consistent training.

To explore the influence of pre-training data on self-consistent training, we no longer add it in each round when training the discriminator, and only the generated data is used. But when the generator is trained, pre-training data is still retained to prevent language degeneration and lack of expressive diversity of the generation. The result of removing pre-training data is shown as the green curves in Figure~\ref{ablation}. With all other training parameters being the same, after the same number of training rounds, the discriminator is slightly worse compared to the original method (red curves in Figure~\ref{ablation}). However, the green curves maintain an upward trend and are very close to the red curves in all datasets except CHIP-STS. This shows that the generated data plays a key role in continuously improving the discriminator, while the pre-training data has a limited role. 

In order to explore the effect of the selection mechanism on training the discriminator, we remove the selection mechanism when training the discriminator, while the training of the generator remains unchanged. The blue curves in Figure~\ref{ablation} depict the performance of the discriminator in each round after removing the selection mechanism. Compared to the original method (red curves), the discriminator only improves in the first round after removing the selection mechanism, which demonstrates the importance of the selection mechanism on the discriminator for the convergence of the self-consistent learning framework. 

\subsection{Experiments on Different Threshold Functions}
To compare the effect of different threshold functions on the final result, we use four type of functions, including oscillatory function (cosine), constant function and monotonically increasing functions (quadratic and linear). For the fairness of comparison, we keep the maxima and minima the same for all functions(except for the constant threshold), and the values are given in supplementary material.

The best results and the second-best results are \textbf{bold} and \underline{underlined}, respectively. As can be seen from the Table~\ref{zero_shot_func}, in the zero-shot setting, the chosen linear function outperforms the other functions, and all the threshold functions show an averaging 10+ AP improvement relative to the baseline. Therefore, the self-consistent learning framework makes it easy to choose a certain threshold function and perform well, and the results are not so sensitive to the choice of the functions. A more detailed figure of the effect of different threshold functions on the results is shown in supplementary material. 

\begin{table}
\small
\centering
\setlength\tabcolsep{4pt} 
\caption{F1 Score of Different Threshold Functions in Zero-Shot.}
\label{zero_shot_func}
\begin{tabular}{l|cccccc}
\toprule
            &  AFQMC          &  CHIP-STS       &  QQP-ZH    &  MRPC           &  AVG            \\ \midrule
Baseline    & 38.25          & 58.82          & 57.88          & 68.54          & 55.87          \\
Cosine & 47.38          & \uline{74.26}    & 64.39          & 83.48    & 67.38          \\
Constant       & 47.06 & 74.15          & 68.67 & \uline{84.11}    & 68.50          \\
Quadratic      & \textbf{51.75} & 73.09          & \textbf{70.85} & 83.48          & 69.79          \\ \midrule
Linear  & \uline{51.36}    & \textbf{77.42} & \uline{70.51}    & \textbf{84.61} & \textbf{70.98} \\ \bottomrule
\end{tabular}
\end{table}

\begin{table}
\small
\centering
\setlength\tabcolsep{4pt} 
\caption{F1 Score of Different Threshold Functions in Fine-Tune.}
\label{fine_tune_func}
\begin{tabular}{l|cccccc}
\toprule
            &  AFQMC          &  CHIP-STS       &  QQP-ZH    &  MRPC           &  AVG            \\ \midrule
Baseline    & 64.19          & 86.93          & 76.56          & 91.21          & 79.72          \\
Cosine & 66.43          & 88.01          & 77.33          & 92.63          & 81.10          \\
Constant  & \uline{66.57}    & \uline{88.15}    & 78.45          & 92.51        & 81.42          \\
Quadratic      & 66.37          & 87.76          & \textbf{79.26} & \uline{92.75}    & 81.54          \\ \midrule
Linear  & \textbf{66.59} & \textbf{88.39} & \uline{78.43}    & \textbf{92.78} & \textbf{81.55} \\ \bottomrule
\end{tabular}
\end{table}

Table~\ref{fine_tune_func} shows the effects of different threshold functions in the fine-tune experiment. It can be seen that all functions have a $1\sim 2$ AP increase relative to the baseline, and the chosen linear function achieves the best performance on all datasets except QQP-ZH.

\subsection{Contrast Experiments with Adversarial Training}

We further demonstrate the superiority of the cooperative approach by comparing the results with adversarial experiments. All experimental settings independent of the training method remain the same in the adversarial training. 

During the experiments, the generator is no longer trained using the samples filtered by the discriminator, but the rewards passed by the discriminator assist the training. All generated samples are treated as negative samples when training the discriminator.

Specifically, $\mathcal{G}$ takes the prompt ' "$s^a$" is similar to " ' and the first $M$ tokens of $s^b$ as input to get $M$ sentence pairs $<s^a, s^b_m>$, where $m$ is from 1 to $M$. Note that we repeat the process of generating sentences $N$ times to reduce the negative impact caused by the large variance of the rewards.\footnote{In practice, we take $M=5,N=5$ for ease of calculation.} The sentence pair is formalized as

\begin{equation}
\nonumber
<s^a, s^b_m>=\mathcal{G}_\theta(s^b_m|s^b_{<m}, \boldsymbol{s^a} ; N)
\end{equation}

Once the $M*N$ sentence pairs are generated, they are passed as input to the $\mathcal{D}$ to obtain the probability score $Q_m^n$ for each of them. We take the average of $Q_m^n$ over $N$ as the reward $\bar Q_m$ corresponding to the $m$-th token. If the sentence length of $s^b$ is greater than $M$, the rewards of the remaining tokens are all the same as those of the $M$-th token. Taking the $m$-th token as an example, the rewards $\bar Q_m$ can be formalized as

\begin{equation}
\nonumber
\begin{array}{l}
\bar Q_{\mathcal{D}_\phi}^{\mathcal{G}_\theta}(m) =\left\{\begin{array}{cc}
\frac{1}{N} \sum_{n=1}^{N} \mathcal{D}_\phi(g_m^n) & m \leq M \\
\bar Q(M) & m>M
\end{array}\right.
\end{array}
\end{equation}

where $g_m^n$ is the $n$-th sentence pair with length $m$.\looseness=-1

Therefore, the objective function for training the generator $\mathcal{G}$ is,

\begin{equation}
\nonumber
	\mathcal{L}_\mathcal{G}(\boldsymbol{s^a}, \boldsymbol{s^b}) = -\frac{1}{|\boldsymbol{s^b}|} \sum_{t=1}^{|\boldsymbol{s^b}|}\log(p_{\mathcal{G}}(s^b_t|s^b_{<t}, \boldsymbol{s^a}) * \bar Q_t)
\end{equation}

The loss function of training the discriminator remains the same as Equation \ref{train_dis}, but differing from cooperative training, the generated samples are regarded as negative samples to the discriminator, and the training target for the discriminator can be given by

\begin{equation}
\nonumber
	\min_{\phi} -\mathbb{E}_{X \sim p_{\text {data}}}\left[\log \mathcal{D}_\phi(X)\right] -\mathbb{E}_{X \sim p_{\mathcal{G}_\theta}}\left[\log \left(1-\mathcal{D}_\phi(X)\right)\right]
\end{equation}

The results of zero-shot and fine-tune on the four datasets are shown in Tables~\ref{zero_shot_adv} and \ref{fine_tune_adv}.

\begin{table}
\small
\centering
\setlength\tabcolsep{6pt} 
\caption{F1 Score of Adversarially Trained Discriminator in Zero-Shot Setting.}
\label{zero_shot_adv}
\begin{tabular}{c|cccc}
\toprule
Round & AFQMC & CHIP-STS & QQP-ZH & MRPC  \\ \midrule
0 & \uline{38.25} & \uline{58.82} & \uline{57.88} & \uline{68.54} \\ 
1 & 0.0 & 8.73 & 21.71 & 4.19 \\
2 & 0.02 & 7.13 & 49.30 & 7.06 \\
3 & 0.0 & 0.29 & 42.94 & 5.32 \\ 
4 & 0.0 & 1.09 & 41.13 & 0.0 \\ 
5 & 0.0 & 0.10 & 43.10 & 1.72 \\ 
6 & 0.0 & 0.39 & 34.30 & 67.38 \\ 
7 & 0.0 & 0.20 & 42.62 & 48.31 \\
8 & 0.0 & 0.20 & 34.95 & 37.97 \\ 
9 & - & 0.20 & 41.81 & - \\ 
10 & - & 0.20 & 40.00 & - \\ \bottomrule
\end{tabular}
\end{table}

As can be seen from Table~\ref{zero_shot_adv}, in the zero-shot setting, training in an adversarial manner does not give any improvement over the baseline. Because the initial discriminator in the zero-shot setting is very weak in distinguishing positive and negative samples, it is reasonable to believe that if all generated samples are considered negative samples from the very beginning, it is difficult for the discriminator to know how to distinguish positive samples. As a result, the F1 scores on both AFQMC and CHIP-STS datasets end up being 0,  while the scores on the QQP-ZH and MRPC datasets fluctuate intensively with the number of rounds,  which further validates the instability of the adversarial training in the zero-shot setting.

\begin{table}
\small
\centering
\setlength\tabcolsep{2pt} 
\caption{F1 score of the Discriminator in Fine-Tune Setting.}
\label{fine_tune_adv}
\begin{tabular}{l|cccccc}
\toprule
            & AFQMC          & CHIP-STS       & QQP-ZH    & MRPC           & AVG            \\ \midrule
 Baseline    & 64.19          & 86.93          & 76.56          & 91.21          & 79.72          \\
 Adversarial & 58.37          & 80.46          & 77.93          & 92.18          & 77.24          \\
 \begin{tabular}[c]{@{}c@{}}Cooperative\\ (\textsc{Our Method})\end{tabular}  & \textbf{66.59} & \textbf{88.39} & \textbf{78.43}    & \textbf{92.78} & \textbf{81.55} \\ \bottomrule
\end{tabular}
\end{table}

For the fine-tune experiments, Table~\ref{fine_tune_adv} shows that training in an adversarial manner can slightly improve the performance on the QQP-ZH and MRPC datasets, but is still worse than the cooperative training. On the AFQMC and CHIP-STS dataset, adversarial training makes it even worse relative to the baseline. It is worth noting that the whole process of adversarial training is so unstable and it is easy to collapse after a few training rounds.

\section{Conclusion}
In this paper, we propose a self-consistent learning framework in the text field to enable cooperative training of the generator and the discriminator.  During the training process, the generator and the discriminator continuously enhance each other until reaching a score consensus. This framework can utilize both limited labeled data and large-scale unlabeled domain-related corpus. Experimental results on four Chinese / English datasets demonstrate that as a form of closed-loop training, our proposed framework can outperforms the strong baselines with continuously improved generators and discriminators. 


\section{Acknowledgements}
This research is supported by National Natural Science Foundation of China (Grant No.62276154), Research Center for Computer Network (Shenzhen) Ministry of Education, Beijing Academy of Artificial Intelligence (BAAI), the Natural Science Foundation of Guangdong Province (Grant No. 2023A1515012914), Basic Research Fund of Shenzhen City (Grant No. JCYJ20210324120012033 and JSGG20210802154402007), the Major Key Project of PCL for Experiments and Applications (PCL2021A06), and Overseas Cooperation Research Fund of Tsinghua Shenzhen International Graduate School (HW2021008).

\bibliography{ecai}
\bibliographystyle{unsrt}


\end{document}